\documentclass{article}

\usepackage{microtype}
\usepackage{graphicx}
\usepackage{subcaption}
\usepackage{booktabs} 
\usepackage{url}            
\usepackage{algorithm}
\usepackage{algorithmic}
\usepackage[table]{xcolor}
\usepackage{multirow}
\usepackage{makecell}
\usepackage{hyperref}
\usepackage{listings}
\usepackage{xcolor}
\usepackage{courier}
\usepackage{siunitx}

\makeatletter
\newcommand{\removelatexerror}{\let\@latex@error\@gobble}
\makeatother

\newcommand{\eat}[1]{}

\long\def\comment#1{}

\renewcommand{\algorithmiccomment}[1]{\hfill $\triangleright$ \textit{#1}}

\definecolor{myblue}{HTML}{324661}
\definecolor{myred}{HTML}{AA6F71}
\definecolor{myyellow}{HTML}{c2ac49}
\definecolor{dartblue}{RGB}{220,245,245}
\definecolor{gain}{RGB}{34,139,34}

\usepackage[preprint]{icml2026}

\usepackage{amsmath}
\usepackage{amssymb}
\usepackage{mathtools}
\usepackage{amsthm}

\usepackage[capitalize,noabbrev]{cleveref}

\newcommand{\ours}{DART}

\usepackage[textsize=tiny]{todonotes}

\icmltitlerunning{DART: Diffusion-Inspired Speculative Decoding for Fast LLM Inference}

\begin{document}

\twocolumn[
  \icmltitle{
    \texorpdfstring{
      \raisebox{-0.25\height}{\includegraphics[width=0.04\textwidth]{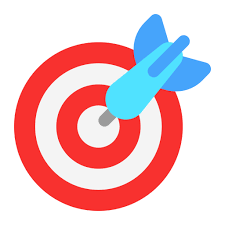}}
      DART: Diffusion-Inspired Speculative Decoding for Fast LLM Inference
    }{
      DART: Diffusion-Inspired Speculative Decoding for Fast LLM Inference
    }
  }



  \icmlsetsymbol{equal}{*}

  \begin{icmlauthorlist}
    \icmlauthor{Fuliang Liu}{nju,alibaba}
    \icmlauthor{Xue Li}{alibaba}
    \icmlauthor{Ketai Zhao}{nju}
    \icmlauthor{Yinxi Gao}{nju}
    \icmlauthor{Ziyan Zhou}{nju}
    \icmlauthor{Zhonghui Zhang}{nju}
    \icmlauthor{Zhibin Wang}{nju}
    \icmlauthor{Wanchun Dou}{nju}
    \icmlauthor{Sheng Zhong}{nju}
    \icmlauthor{Chen Tian}{nju}
  \end{icmlauthorlist}

  \icmlaffiliation{nju}{State Key Laboratory of Novel Software Technology, Nanjing University}
  \icmlaffiliation{alibaba}{Alibaba Group}

  \icmlcorrespondingauthor{Zhibin Wang}{wzbwangzhibin@gmail.com}

  \icmlkeywords{Machine Learning, ICML}

  \vskip 0.3in
]



\printAffiliationsAndNotice{}  

\begin{abstract}
  Speculative decoding is an effective and lossless approach for accelerating LLM inference. However, existing widely adopted model-based draft designs, such as EAGLE3, improve accuracy at the cost of multi-step autoregressive inference, resulting in high drafting latency and ultimately rendering the drafting stage itself a performance bottleneck. Inspired by diffusion-based large language models (dLLMs), we propose DART, which leverages parallel generation to reduce drafting latency. DART predicts logits for multiple future masked positions in parallel within a single forward pass based on hidden states of the target model, thereby eliminating autoregressive rollouts in the draft model while preserving a lightweight design. Based on these parallel logit predictions, we further introduce an efficient tree pruning algorithm that constructs high-quality draft token trees with \textit{N}-gram–enforced semantic continuity. DART substantially reduces draft-stage overhead while preserving high draft accuracy, leading to significantly improved end-to-end decoding speed. Experimental results demonstrate that DART achieves a $2.03\times$--$3.44\times$ wall-clock time speedup across multiple datasets, surpassing EAGLE3 by 30\% on average and offering a practical speculative decoding framework. Code is released at \href{https://github.com/fvliang/DART}{https://github.com/fvliang/DART}.
\end{abstract}

\section{Introduction}

\begin{figure}[t]
  \centering
  \includegraphics[width=\columnwidth]{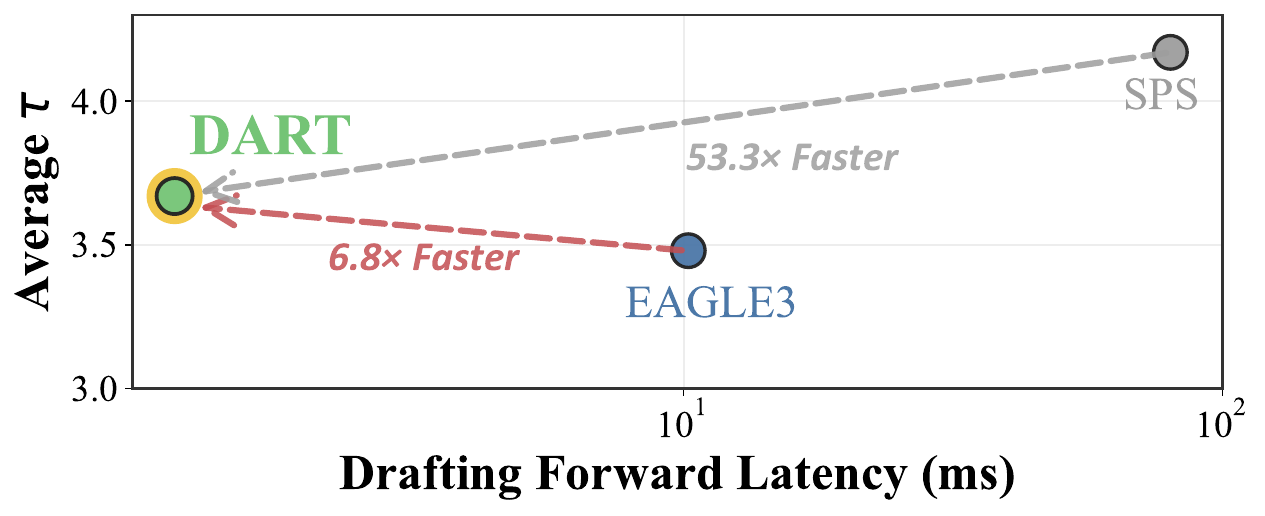}
  \caption{\textit{Average Acceptance Length} ($\tau$) versus drafting forward latency (ms), averaged across all benchmarks, for speculative decoding of Qwen3-14B on H20-3e GPU. Compared with EAGLE3 and SPS (standard speculative sampling; using Qwen3-1.7B as the draft model with a draft length of 5, set to yield a measurable speedup, whereas both EAGLE3 and DART use a draft length of 8), DART reduces drafting forward latency by up to $6.8\times$ and $53.3\times$, respectively, while preserving relatively high $\tau$, demonstrating a significantly improved drafting efficiency.}
  \label{fig:1}
\end{figure}

Speculative decoding \cite{leviathan2023fast,chen2023accelerating,sun2023spectr,sun2024optimal,sun2024triforce,gao2025falcon} has emerged as a promising approach to accelerate memory-bound LLM inference, particularly as modern state-of-the-art models scale to hundreds of billions of parameters \cite{guo2025deepseek,liu2024deepseek}. In speculative decoding, a lightweight draft model proposes multiple future tokens, which are then verified by the target model to ensure that the final output distribution exactly matches that of standard autoregressive decoding. Through speculative decoding, multiple tokens could be generated in every iteration to significantly improve inference efficiency. The effectiveness of speculative decoding depends critically on the design of the drafter. First, the drafter should achieve high predictive accuracy, most commonly quantified by the \textit{Average Acceptance Length} ($\tau$), i.e., the average number of accepted draft tokens during verification. A higher $\tau$ directly translates to fewer target-model invocations and greater decoding speedups. Second, the draft model itself must be computationally efficient, which could be quantified as drafting latency $(ms)$, as excessive drafting overhead can significantly erode the overall benefits of speculative decoding.

\begin{figure*}[t]
  \centering
  \includegraphics[width=\textwidth]{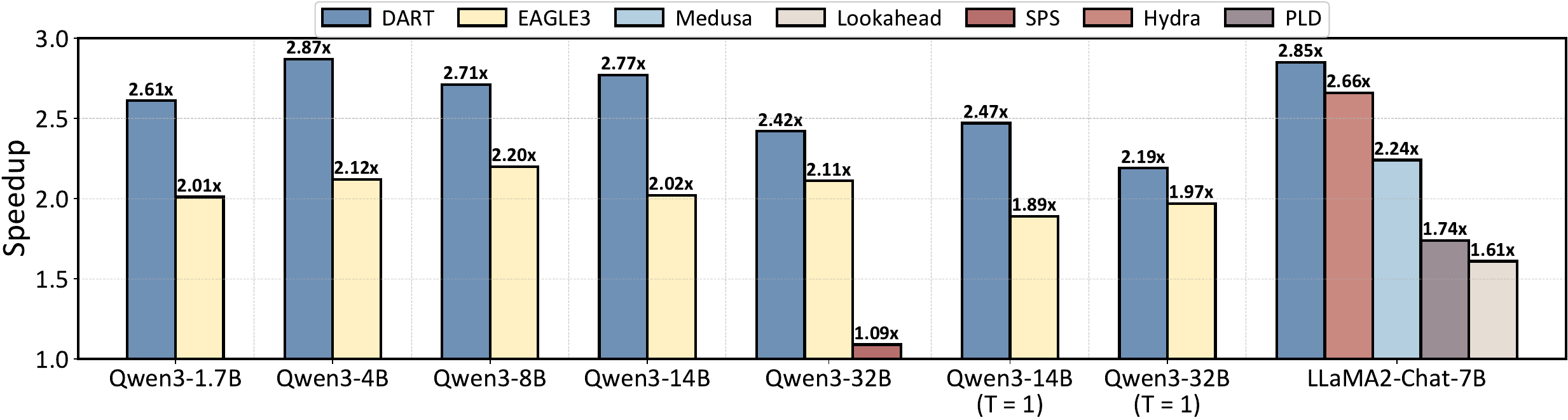}
  \caption{Speedup over vanilla autoregressive decoding (batch size $=1$), averaged across all datasets. For Qwen3 models, results are reported at temperature $T=0$ on Qwen3-{1.7B, 4B, 8B, 14B, 32B}, and additionally at $T=1$ on Qwen3-{14B, 32B}; only DART and EAGLE3 are compared on Qwen3, except for Qwen3-32B at $T=0$, where we also include SPS (using Qwen3-1.7B as the draft model with draft length 5).
  For LLaMA2-Chat-7B, we compare DART with methods including Medusa, Lookahead, SPS, and PLD.}
  \label{fig:2}
\end{figure*}

Typical speculative decoding's drafter implementation is autoregressive. In vanilla speculative decoding \cite{leviathan2023fast}, the drafter is typically a lower-parameter variant from the same model family as the target model. However, vanilla speculative decoding often suffers from high drafting latency, which can account for over 75\% and 60\% of the total inference time when using Qwen3-1.7B to accelerate Qwen3-14B and Qwen3-32B with draft length 5. In pursuit of lower draft latency, Medusa \cite{cai2024medusa} applies lightweight decoding heads to predict multiple subsequent tokens on the top-layer features of the target model but delivers limited accuracy. EAGLE \cite{li2024eagle, li2024eagle2} improves the accuracy by customizing a dedicated single layer to predict next-feature autoregressively. Subsequent methods \cite{li2025eagle3,zhang2025learning} such as EAGLE3 further boost accuracy through reducing inconsistency between training and decoding of draft model.

However, current autoregressive drafter designs introduce limitations. While approaches such as EAGLE3 significantly lower the per-step drafting cost by using a single customized layer, the drafting process is still inherently autoregressive. This sequential dependency forces the drafter to spend nearly 20\%--40\% of the total inference time, thereby fundamentally limiting the achievable acceleration, causing the drafting stage, especially the drafting forward cost, to emerge as a new bottleneck in speculative decoding. At the other extreme, some speculative decoding approaches, such as Lookahead \cite{fu2024break}, rely solely on \textit{N}-gram or retrieval-based heuristics for drafting. While such methods incur negligible drafting latency, their limited predictive accuracy typically leads to very low average acceptance length $\tau$, resulting in modest end-to-end speedups.

Using diffusion-style parallel generation appears to be a promising direction for overcoming the above limitations. However, directly adopting diffusion-based large language models (dLLMs) as draft models \cite{christopher2025speculative, sandler2025specdiff, li2025diffuspec, cheng2025deer} suffers from fundamental limitations. For example, DiffuSpec \cite{li2025diffuspec} uses Dream7B \cite{ye2025dream} as the draft model to accelerate Qwen3-32B and \textit{N}-gram based CPS to give a draft sequence in single forward. However, existing dLLMs like Dream7B are designed for bidirectional context modeling, where the objective is to jointly model and refine tokens across an entire sequence with full contextual access. In contrast, speculative decoding for LLM inference requires predicting a contiguous span of future tokens conditioned strictly on a given prefix, preserving the causal structure of autoregressive generation. This fundamental mismatch in modeling assumptions makes dLLMs inherently unsuitable as drop-in draft models for speculative decoding. Moreover, although dLLMs enable parallel token generation, directly using them as draft models does not translate into lower drafting latency in practice. Because dLLMs such as Dream7B are instantiated as standalone models with substantial parameter, their per-step inference cost remains significantly higher than lightweight, target-coupled drafters such as EAGLE3—often by tens of times in practice. As a result, the overall drafting latency of dLLM-based approaches is often substantially higher, despite their parallel decoding capability. More critically, directly using dLLMs introduces additional practical issues, including tokenizer incompatibility and limited availability.

In this paper, we design a customized diffusion-style drafter tailored for speculative decoding and propose \ours{}, a diffusion-inspired drafting approach.

\begin{figure*}[t]
  \centering
  \includegraphics[width=\textwidth]{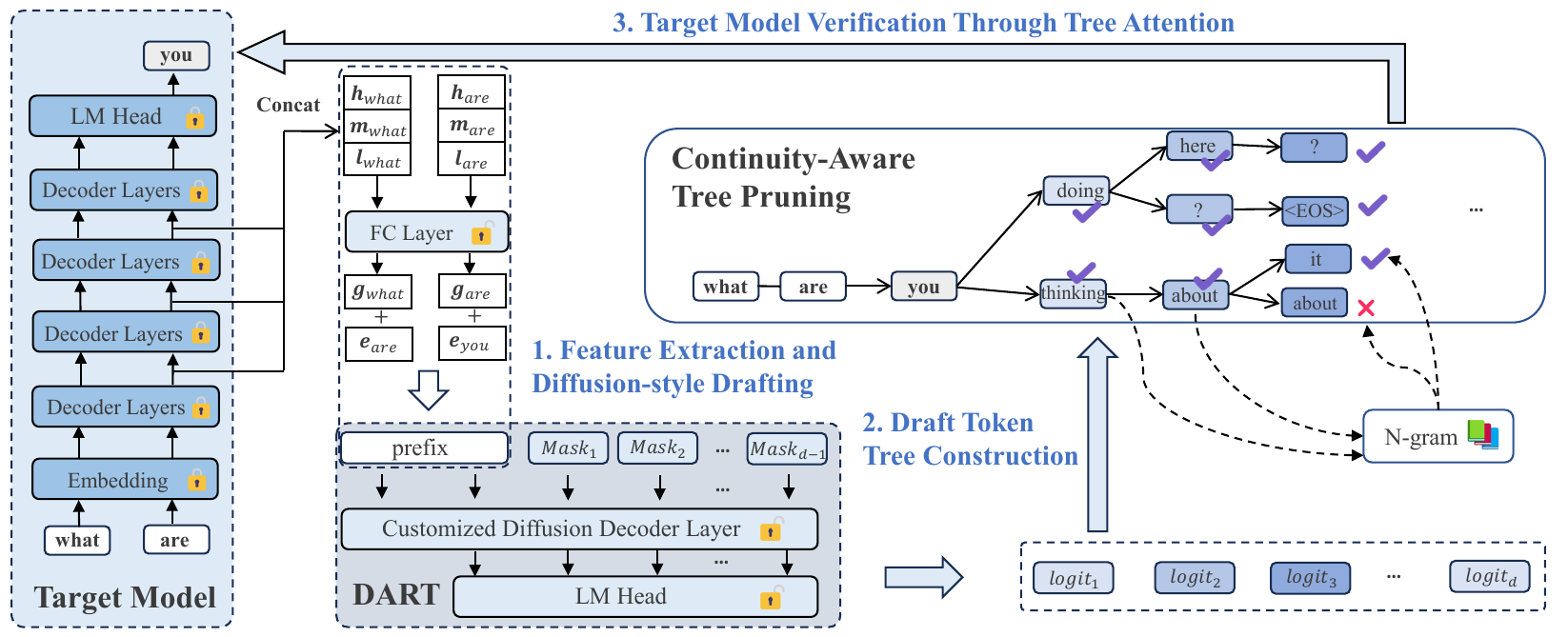}
  \caption{Diagram of the DART inference pipeline, illustrating the three substeps of DART's speculative decoding. $\mathbf{l}, \mathbf{m}, \mathbf{h}$ represent the low, middle, and high-level features of the target model, respectively. $\mathbf{e}$ denotes the embedding. Unlocked icon means learnable parameter. After feature extraction, we append $(d-1)$ Mask tokens to the prefix and conduct single forward to get $d$ logits, where the first logit comes from the output of the last position in prefix. We call this ``Shifted logits prediction'' in Section~\ref{sec:dart_draft}. During ``Continuity-Aware Tree Pruning'', candidate tokens are selected from the corresponding position's predicted logit and a \textit{N}-gram model ensures that the expanded tokens maintain continuity. After getting the final pruned draft token tree, we verify them in target model with Tree Attention.}
  \label{fig:3}
\end{figure*}

\ours{} is designed according to the following two principles. \textbf{First, diffusion-style parallel drafting is only beneficial when explicitly customized for speculative decoding.} Rather than naively adopting conventional dLLMs with bidirectional context modeling, the draft model should be tailored to predict the next few positions conditioned on a given prefix, aligning with the causal requirements of speculative decoding while retaining parallel generation. In this paper, \ours{} adopts an efficient customized training recipe that enables the draft model to learn the distribution of multiple future positions conditioned on the given prefix. \textbf{Second, effective drafting should preserve the low-cost design.} To keep the drafting overhead negligible, \ours{} is tightly coupled with the target model: it reuses the target model’s hidden states as input and applies only a single lightweight layer to directly produce logits for multiple future positions, instead of adopting a drop-in dLLM with prohibitive inference latency. However, these predicted logits implicitly induce an exponentially large combinatorial space of possible token continuations, \ours{} further employs an efficient tree pruning algorithm to construct the final draft token tree for verification, achieving low latency while maintaining high-quality candidates with \textit{N}-gram.

In summary, we propose \ours{}, a diffusion-inspired speculative decoding framework that rethinks the design of draft models under the constraints of drafting latency. Our main contributions are summarized as follows:

\begin{itemize}
  \item \textbf{Lightweight diffusion-style parallel drafting.}
        We are the first to introduce a draft model that operates directly on the target model’s hidden states and predicts multiple future logits in parallel using a single customized layer. This design completely eliminates autoregressive rollout in the drafter and removes the need for complex KV cache management of autoregressive draft model. As shown in Figure~\ref{fig:1}, \ours{} achieves up to \textit{6.8$\times$ faster} drafting forward than autoregressive drafters such as EAGLE3, while remaining relatively high average acceptance length $\tau$.

  \item \textbf{Continuity-aware tree pruning via \textit{N}-gram.}
        We identify that parallel logits prediction induces an exponentially large combinatorial search space, which cannot be efficiently handled by naive decoding. Instead of using \textit{N}-gram models as standalone draft predictors, which suffer from low acceptance rates, we redesign \textit{N}-gram as a continuity-aware pruning mechanism to efficiently constrain the draft token tree constructed from parallel logits. This continuity-aware pruning preserves high-quality candidates and substantially improves the average acceptance length $\tau$.

 \item \textbf{Significant end-to-end acceleration.}
        As shown in Figure~\ref{fig:2}, extensive experiments across multiple benchmarks demonstrate that \ours{} achieves substantial end-to-end throughput improvements, with up to \textit{2.03--3.44$\times$} speedup over standard autoregressive decoding. Compared to prior speculative decoding methods such as EAGLE3, \ours{} delivers around \textit{30\%} higher speedup on average, while further achieving up to \textit{65\%} improvement on certain code-centric workloads under the same target model setting. These results validate a new speculative decoding paradigm that combines low drafting latency and high average acceptance length.

\end{itemize}

\section{Preliminaries}

\subsection{Speculative Decoding}

Let $q_\theta$ denote the target autoregressive language model and $p_\phi$ a lightweight draft model.
Standard autoregressive decoding generates one token per forward pass of $q_\theta$, which incurs high inference latency. Speculative decoding accelerates generation by allowing the draft model to first propose a block of $K$ future tokens conditioned on the current prefix.
The target model then verifies these draft tokens in parallel and accepts a contiguous prefix of them using a rejection-based procedure, guaranteeing that the final output distribution exactly matches $q_\theta$. The effectiveness of speculative decoding is commonly measured by the expected number of accepted draft tokens $\tau$, which determines the amortized reduction in target-model invocations. Practical draft models therefore aim to maximize $\tau$ while keeping the drafting overhead minimal.

\subsection{Diffusion-Based Large Language Models}

Diffusion-based large language models (dLLMs) \cite{nie2025large,ye2025dream,khanna2025mercury} generate text by iteratively denoising partially masked sequences, predicting multiple token positions in parallel conditioned on the surrounding context. This masked, non-autoregressive formulation enables holistic sequence modeling and significantly reduces generation steps compared to autoregressive decoding. The strong parallel prediction capability of dLLMs motivates our design in DART, where we seek to eliminate autoregressive rollout in the drafting stage of speculative decoding. However, standard dLLMs are inherently bidirectional and operate on full sequences, which is incompatible with the strictly prefix-conditioned requirement and exactness guarantees of speculative decoding. DART therefore adopts a diffusion-inspired masked prediction mechanism that is explicitly tailored to speculative decoding, rather than performing full-sequence denoising.

\section{DART}
\label{sec:dart}

In this section, we will go through the details of DART.

\subsection{Requirements of Speculative Decoding}

The draft model in speculative decoding for autoregressive LLMs is subject to several unique requirements, which fundamentally distinguish \ours{} from standard diffusion-based text generation:

\begin{itemize}
  \item \textbf{Causal attention mask.}
  Since all prefix tokens are fully accessible and all future positions are predicted in a single forward pass, bidirectional attention for iterative token refinement is unnecessary. \ours{} therefore retains a strictly causal attention mask, both over the prefix and within the masked block.

  \item \textbf{Limited drafting horizon.}
  Due to the inevitable distribution mismatch between the lightweight draft model and the target model, speculative decoding typically benefits from predicting only a small number of future tokens (e.g., 8). Since \ours{} is designed to generate only a limited number of draft tokens, it can achieve relatively high accuracy even without iterative bidirectional attention refinement.

  \item \textbf{Positional importance bias.}
  Earlier draft positions are substantially more important than later ones, as token acceptance proceeds sequentially from the prefix--an error in an early token immediately terminates the verification process.
\end{itemize}

These characteristics impose stringent constraints on the design of efficient diffusion-style draft models and motivate specialized architectures that prioritize early-token accuracy, controlled horizon prediction, and strict causal conditioning.

\subsection{Diffusion-Inspired Drafting Phase}
\label{sec:dart_draft}

\paragraph{Draft model architecture.}
DART follows the lightweight design principle established by prior speculative decoding methods such as EAGLE3 \cite{li2025eagle3}. The draft model consists of a single customized Transformer decoder layer that operates on intermediate representations of the target model. Concretely, after a prefilling or verification forward pass of the target model, we extract hidden states from multiple intermediate layers (denoted as $\mathbf{h}, \mathbf{m}, \mathbf{l}$ in Figure~\ref{fig:3}). These hidden states are concatenated and projected through a fully connected layer to obtain a compact prefix representation $\mathbf{g}_{1:n} \in \mathbb{R}^{n \times k}$. In addition, we incorporate shifted token embeddings $\mathbf{e}_{2:n+1}$, obtained by sampling the next token $\mathbf{t}_{n+1}$ from the output of target model and applying the embedding layer of the target model. The final prefix input to the draft model is formed by concatenating $\mathbf{g}_{1:n}$ and $\mathbf{e}_{2:n+1}$ along the feature dimension, which is denoted as $\mathbf{z}_{1:n}$ . To enable parallel prediction of future tokens, DART appends a fixed-length suffix of $d-1$ $\langle\textsc{mask}\rangle$ tokens to the prefix, yielding the following input sequence:
\[
  [\; \mathbf{z}_{1:n},\ \langle\textsc{mask}\rangle_{n+1:n+d-1} \;].
\]
A single forward pass of the draft model yields logits for all future positions simultaneously:
\[
  \{\boldsymbol{\ell}_{n+1}, \boldsymbol{\ell}_{n+2}, \dots, \boldsymbol{\ell}_{n+d}\},
\]
where $\boldsymbol{\ell}_{n+1}$ is read from the output at the last prefix position, and subsequent logits are read from masked positions.

\paragraph{Shifted logits prediction.}
In dLLMs, logits predicted at masked positions are typically interpreted as predictions for the tokens at the same positions. In contrast, \ours{} adopts a \emph{shifted logits prediction} scheme, where the logit produced at each position corresponds to the prediction of the next token. Empirically, we find that this shifted formulation significantly improves the prediction accuracy at the first drafted position. Moreover, it enables more efficient training by allowing all positions to contribute supervision signals.

\begin{figure}[t]
  \centering
  \includegraphics[width=\columnwidth]{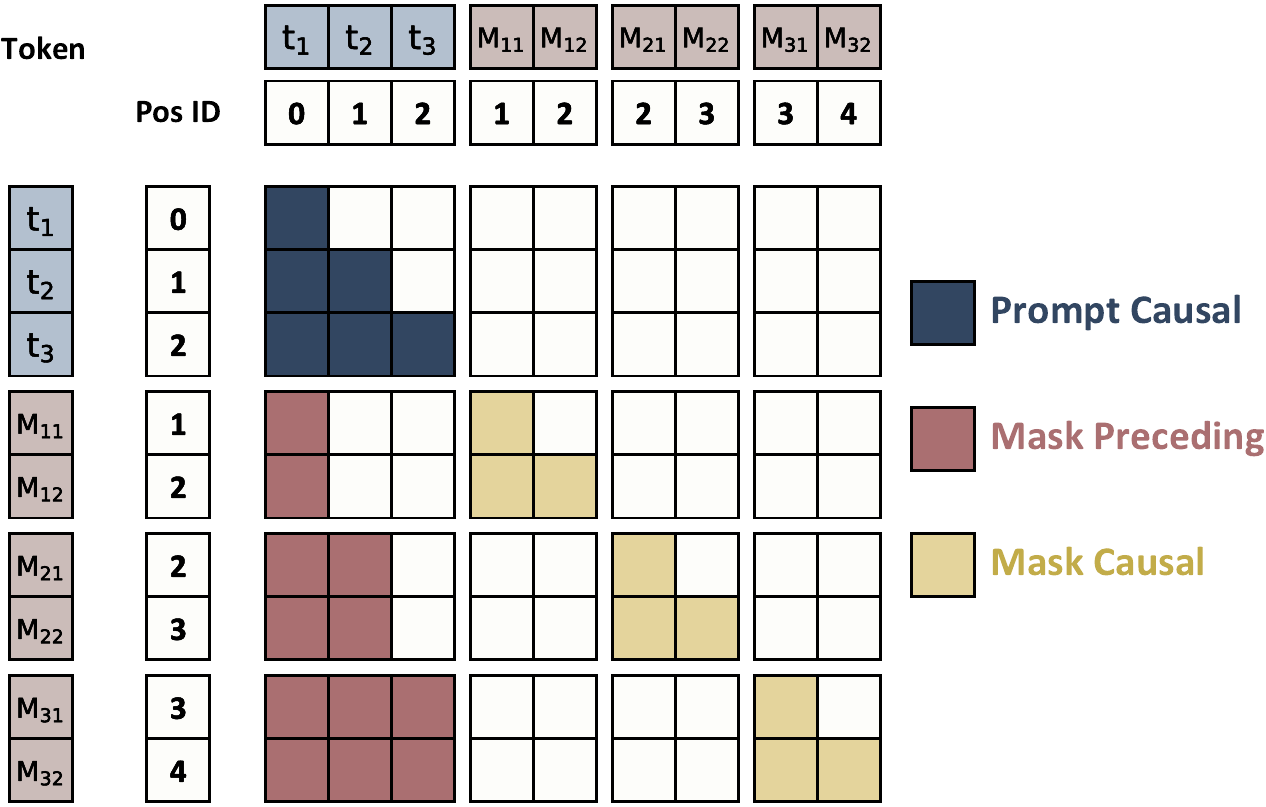}
  \caption{Position IDs and attention mask during prefix-share training. The
    attention mask combines clean data causal attention (\textcolor{myblue}{Prompt Causal}), prefix attention for every mask block (\textcolor{myred}{Mask Preceding}), and block-inner causal attention (\textcolor{myyellow}{Mask Causal}). For clarity of presentation, the figure depicts a simplified example with a prefix length of 3 and mask block length of 2, rather than mask block length of 7 in the actual DART training.}
  \label{fig:4}
\end{figure}

\paragraph{Parallel holistic prediction.}
DART is inspired by dLLMs, which predict multiple masked positions in parallel.
However, DART does \emph{not} perform iterative denoising or bidirectional refinement. Instead, it directly predicts the conditional token distributions of multiple future positions \emph{in a single step}, strictly conditioned on the given prefix. Through holistic prediction, \textbf{sequential latency is eliminated}: the drafting cost is independent of the draft length $d$, requiring only a single forward pass of the lightweight draft model. 

\subsection{Training of the Draft Model}
\label{sec:dart_training}

\paragraph{Prefix-shared masked training.}
To closely match the inference-time drafting behavior, the draft model is trained using a
\emph{prefix-shared, multi-token prediction objective}.
Given a training sequence $x_{1:L}$, we simultaneously construct multiple training instances
at different prefix positions within the same sequence.
For each prefix position $n$, the model is trained to predict the next $d$ tokens
$\{x_{n+1}, \dots, x_{n+d}\}$ under a shifted prediction scheme, where supervision is applied
to both the prefix and masked positions, and all predictions are conditioned only on the
prefix $x_{1:n}$.

All such prefix positions are trained jointly within a single forward pass by employing a carefully designed sparse attention mask.
Specifically, as illustrated in Figure~\ref{fig:4}, masked positions corresponding to different prefixes are allowed to attend exclusively to their respective prefixes. Attention among masked positions within the same block is causal to fit the attention mask during inference, while attention across different blocks is explicitly disabled. This \emph{prefix-isolated} attention structure ensures that each masked position learns to model $p_\phi(\cdot \mid x_{1:n})$, without access to future ground-truth tokens. This design enables efficient supervision of multiple future positions across the sequence in parallel, substantially increasing training efficiency while strictly preserving causal conditioning. In practice, the resulting attention pattern is highly sparse. We further leverage \textit{Flex-Attention} \cite{dong2024flex} to exploit this sparsity, significantly reducing memory consumption and accelerating training without altering the underlying model computation. We give the pseudo \textit{Flex-Attention} code of DART’s sparse attention mask in Appendix \ref{sec:appendix_train}.

\paragraph{Annealed KL divergence objective.}
Instead of supervising the draft model with discrete one-hot targets, DART optimizes a position-aware KL divergence objective between the draft model predictions and the target model distributions.
For each future position $t \in \{1, \dots, d\}$, we minimize
\[
  \mathcal{L}_{\text{KL}} =
  \sum_{t=1}^{d}
  \lambda_t \,
  \mathrm{KL}\!\left(
  q_\theta(\cdot \mid x_{1:n+t-1})
  \;\|\;
  p_\phi(\cdot \mid x_{1:n}, t)
  \right),
\]
where $\lambda_t = \gamma^{\,t-1},$ $p_\phi(\cdot \mid x_{1:n}, t)$ denotes the draft model’s predicted distribution for the $t$-th future position.
This exponentially decaying weighting reflects the increasing uncertainty of longer-horizon predictions and prevents later, noisier targets from dominating the training signal.
In practice, we set $\gamma=0.6$, which consistently yields the best trade-off between early-position accuracy and overall training stability in our ablation studies (Section~\ref{sec:ablation}).
By emphasizing short-horizon predictions while softly regularizing distant ones, this annealed objective substantially improves draft quality at the most critical positions and thereby provides higher-quality candidate distributions for downstream draft tree construction.

\subsection{What Does DART Output?}
\label{sec:dart_output}

Rather than directly producing a draft token tree, DART outputs a set of parallel logits $\{\boldsymbol{\ell}_{n+1}, \dots, \boldsymbol{\ell}_{n+d}\}$, one for each future position.
Each logit encodes a high-quality candidate set for that position conditioning on the prefix.
Together, these logits define a compact but expressive \emph{candidate set} that implicitly
contains many plausible future continuations.

\section{Efficient \textit{N}-gram-based Tree Pruning}
\label{sec:dart_tree}

\subsection{Implicit Huge Search Space in Parallel Logits}
The $d$ parallel logits predicted by DART define a factorized distribution over future positions, which implicitly induce an exponentially large combinatorial space of possible token continuations, corresponding to a full $d$-level token tree. Directly verifying this space is infeasible, and naively combining independently predicted tokens may lead to locally semantically implausible draft sequences. The goal of draft tree construction is therefore to extract a compact, high-quality subset of this implicit tree that can be efficiently verified by the target model.

\begin{algorithm}[t]
  \caption{Continuity-Aware Tree Pruning}
  \label{alg:pipeline}
  \small
  \begin{algorithmic}[1]
    \INPUT $d$ future-position logits $\{l_i\}_{i=1}^d$;
    \textit{N}-gram model $g_n$;
    candidate threshold $\{k_i\}_{i=1}^d$;
    prefix context $ctx$;
    tree size $\theta$;
    beam width $w$
    \OUTPUT Final draft token tree $\mathcal{T}$

    \STATE $\mathcal{C} \gets \{(ctx, 0)\}$ \algorithmiccomment{Active candidate set (sequence, score)}
    \STATE $\mathcal{T} \gets \varnothing$ \algorithmiccomment{Global token tree}

    \FOR{$i = 1$ to $d$}
    \STATE $\mathcal{C}' \gets \varnothing$
    \FORALL{$(seq, sc) \in \mathcal{C}$ \textbf{in parallel}}
    \STATE $\mathcal{S}_i \gets \text{Top-}k_i\text{ tokens from } l_i$
    \FORALL{$t \in \mathcal{S}_i$}
    \STATE $s_{\text{logit}} \gets \log(softmax(l)_{i,t} + \epsilon)$
    \STATE $s_{\text{ng}} \gets g_n(t, seq_{-n:})$ \algorithmiccomment{\textit{N}-gram continuity score}
    \STATE $s' \gets sc + \textsc{Combine}(s_{\text{logit}}, s_{\text{ng}})$
    \STATE $\mathcal{C}' \gets \mathcal{C}' \cup \{(\textsc{Cat}(seq, t), s')\}$
    \STATE $\mathcal{T} \gets \mathcal{T} \cup \{(\textsc{Cat}(seq, t), s')\}$
    \ENDFOR
    \ENDFOR
    \STATE $\mathcal{C} \gets \text{Top-}w \text{ scoring pairs from } \mathcal{C}'$
    \ENDFOR

    \STATE \textbf{return} Top-$\theta$ scoring nodes from $\mathcal{T}$
  \end{algorithmic}
\end{algorithm}

\subsection{Consistency-constrained Pruning via \textit{N}-gram}
To extract a compact and semantically coherent draft tree from the exponentially large search space induced by parallel logits, DART performs consistency-constrained pruning guided by an established \textit{N}-gram model, as shown in Algorithm \ref{alg:pipeline}. Starting from the prefix context, candidate tokens at each future position are first locally ranked according to their draft logits. During tree expansion, these candidates are further scored using an \textit{N}-gram continuity score that measures the likelihood of appending a token given the recent $n$-token suffix of the partial sequence. The logit-based score and the \textit{N}-gram score are combined to form a unified ranking criterion, which favors token sequences that are both locally probable under the draft model and globally consistent with surface-level language statistics. This constraint effectively filters out semantically implausible combinations that arise from independently predicted positions, while retaining high-quality candidates. This design is model-agnostic, decoupling the draft model’s forward computation from the tree expansion procedure. The detailed hyperparameter setting is given in Appendix~\ref{sec:appendix_algr}.

\begin{table*}[ht]
  \centering
  \caption{Speedup ratios of different methods and mean average acceptance lengths $\tau$ on MT-Bench, HumanEval, Alpaca, Math500, CodeAlpaca, LiveCodeBench and MBPP. All Qwen models are from Qwen3 family, for example, Qwen32B represents Qwen3-32B. L2 7B represents LLaMA2-Chat-7B. In SPS, we use the Qwen3-1.7B as the drafter of Qwen3-14B and Qwen3-32B with draft length 5.}
  \label{tab:performence}
  \newcommand{\cb}{\cellcolor{dartblue}}
  \resizebox{\linewidth}{!}{
    \begin{tabular}{ccccccccccc}
    \toprule
    \multirow{2}{*}{Model} &
    \multirow{2}{*}{Method} &
    \multicolumn{7}{c}{Speedup} &
    \multicolumn{2}{c}{Mean} \\
    
    \cmidrule(lr){3-9}
    \cmidrule(lr){10-11}
    
    &
    & Alpaca & Codealpaca & Humaneval & LiveCode & Math500 & MBPP & MT-bench
    & Speedup & $\tau$ \\
    
    \midrule

    \multicolumn{10}{c}{Temperature=0} \\
    \midrule
    \multirow{5}{*}{L2 7B}
      & PLD        & 1.18× & 1.75× & 1.84× & 1.94× & 1.59× & 1.59× & 1.57× & 1.74× & 1.92 \\
      & Lookahead  & 1.42× & 1.54× & 1.64× & 1.59× & 1.82× & 1.61× & 1.63× & 1.61× & 1.81 \\
      & Medusa     & 2.13× & 2.24× & 2.22× & 2.31× & 2.29× & 2.35× & 2.12× & 2.24× & 2.68 \\
      & Hydra      & 2.72× & 2.89× & 2.60× & 2.59× & 2.69× & 2.62× & 2.48× & 2.66× & 3.55 \\
      & \cb\textbf{DART}
                    & \cb\textbf{2.95×}
                    & \cb\textbf{3.03×}
                    & \cb\textbf{2.98×}
                    & \cb\textbf{2.81×}
                    & \cb\textbf{2.84×}
                    & \cb\textbf{2.72×}
                    & \cb\textbf{2.61×}
                    & \cb\textbf{2.85×} & \cb\textbf{4.08} \\
    \midrule

    \multirow{2}{*}{Qwen1.7B}
      & EAGLE3     & 1.84× & 1.93× & 1.94× & 2.06× & 2.25× & 2.05× & 1.98× & 2.01× & \textbf{3.80} \\
      & \cb\textbf{DART}
                    & \cb\textbf{2.60×}
                    & \cb\textbf{2.90×}
                    & \cb\textbf{2.79×}
                    & \cb\textbf{2.28×}
                    & \cb\textbf{2.52×}
                    & \cb\textbf{2.64×}
                    & \cb\textbf{2.57×}
                    & \cb\textbf{2.61×} & \cb3.60 \\
    \midrule

    \multirow{2}{*}{Qwen4B}
      & EAGLE3     & 2.15× & 2.21× & 2.20× & 1.95× & 2.06× & 2.08× & 2.17× & 2.12× & 3.54 \\
      & \cb\textbf{DART}
                    & \cb\textbf{2.55×}
                    & \cb\textbf{3.45×}
                    & \cb\textbf{3.25×}
                    & \cb\textbf{2.57×}
                    & \cb\textbf{2.50×}
                    & \cb\textbf{3.06×}
                    & \cb\textbf{2.73×}
                    & \cb\textbf{2.87×} & \cb\textbf{3.87} \\
    \midrule

    \multirow{2}{*}{Qwen8B}
      & EAGLE3     & 2.02× & 2.34× & 2.37× & 2.11× & 2.30× & 2.20× & 2.08× & 2.20× & \textbf{3.72} \\
      & \cb\textbf{DART}
                    & \cb\textbf{2.51×}
                    & \cb\textbf{3.40×}
                    & \cb\textbf{2.80×}
                    & \cb\textbf{2.64×}
                    & \cb\textbf{2.34×}
                    & \cb\textbf{3.09×}
                    & \cb\textbf{2.22×}
                    & \cb\textbf{2.71×} & \cb3.61 \\
    \midrule

    \multirow{3}{*}{Qwen14B}
      & SPS        & 1.05× & 1.07× & 0.96× & 0.94× & 0.96× & 0.97× & 0.92× & 0.98× & \textbf{4.17} \\
      & EAGLE3     & 1.69× & 2.08× & 2.27× & 2.08× & 2.32× & 1.99× & 1.71× & 2.02× & 3.48 \\
      & \cb\textbf{DART}
                    & \cb\textbf{2.73×}
                    & \cb\textbf{3.44×}
                    & \cb\textbf{2.79×}
                    & \cb\textbf{2.69×}
                    & \cb\textbf{2.50×}
                    & \cb\textbf{2.98×}
                    & \cb\textbf{2.20×}
                    & \cb\textbf{2.77×} & \cb3.67 \\
    \midrule

    \multirow{3}{*}{Qwen32B}
      & SPS        & 1.07× & 1.08× & 1.06× & 1.06× & 1.06× & 1.15× & 1.12×  & 1.09× & 3.45 \\
      & EAGLE3     & 1.72× & 2.38× & 2.19× & 2.15× & 2.31× & 2.27× & 1.76× & 2.11× & \textbf{3.85} \\
      & \cb\textbf{DART}
                    & \cb\textbf{2.03×}
                    & \cb\textbf{2.88×}
                    & \cb\textbf{2.37×}
                    & \cb\textbf{2.42×}
                    & \cb\textbf{2.46×}
                    & \cb\textbf{2.56×}
                    & \cb\textbf{2.24×}
                    & \cb\textbf{2.42×} & \cb3.76 \\
    \midrule

    \multicolumn{10}{c}{Temperature=1} \\
    \midrule

    \multirow{2}{*}{Qwen14B}
      & EAGLE3     & 1.62× & 2.01× & 2.17× & 1.92× & 2.10× & 1.81× & 1.61× & 1.89× & 3.38 \\
      & \cb\textbf{DART}
                    & \cb\textbf{2.38×}
                    & \cb\textbf{3.10×}
                    & \cb\textbf{2.48×}
                    & \cb\textbf{2.37×}
                    & \cb\textbf{2.29×}
                    & \cb\textbf{2.71×}
                    & \cb\textbf{1.94×}
                    & \cb\textbf{2.47×} & \cb\textbf{3.61} \\
    \midrule

    \multirow{2}{*}{Qwen32B}
      & EAGLE3     & 1.57× & 2.30× & 2.08× & 1.97× & 2.15× & 2.17× & 1.58× & 1.97× & \textbf{3.67} \\
      & \cb\textbf{DART}
                    & \cb\textbf{1.84×}
                    & \cb\textbf{2.78×}
                    & \cb\textbf{2.19×}
                    & \cb\textbf{2.13×}
                    & \cb\textbf{2.16×}
                    & \cb\textbf{2.40×}
                    & \cb\textbf{1.82×}
                    & \cb\textbf{2.19×} & \cb3.55 \\
    \bottomrule
    \end{tabular}
  }
\end{table*}

\section{Experiments}

\subsection{Experimental Setup}

\paragraph{Hardware.} All training and inference processes are conducted on a server equipped with 8$\times$NVIDIA H20-3e GPUs (141GB), 90 CPU cores, 900GB of RAM and PyTorch 2.8.0.

\paragraph{Target models.} We mainly train DART on the Qwen3 model family \cite{qwen3}, including Qwen3-1.7B, Qwen3-4B, Qwen3-8B, Qwen3-14B and Qwen3-32B. To fairly compare DART with other speculative decoding methods (e.g., Lookahead, PLD \cite{saxena2023prompt}, Medusa and Hydra \cite{ankner2024hydra}), we additionally train DART on LLaMA2-Chat-7B \cite{touvron2023llama}, since these methods are not compatible with the Qwen3 family.

\paragraph{Benchmarks.}
We evaluate DART on a diverse suite of tasks covering instruction following, multi-turn conversation, mathematical reasoning, and code generation: MT-Bench \cite{zheng2023judging}, HumanEval \cite{chen2021codex}, Alpaca \cite{alpaca}, Math500 \cite{Math500}, CodeAlpaca \cite{CodeAlpaca}, LiveCodeBench \cite{jain2024livecodebench}, and MBPP \cite{MBPP}.

\begin{figure}[t]
  \centering
  \includegraphics[width=\columnwidth]{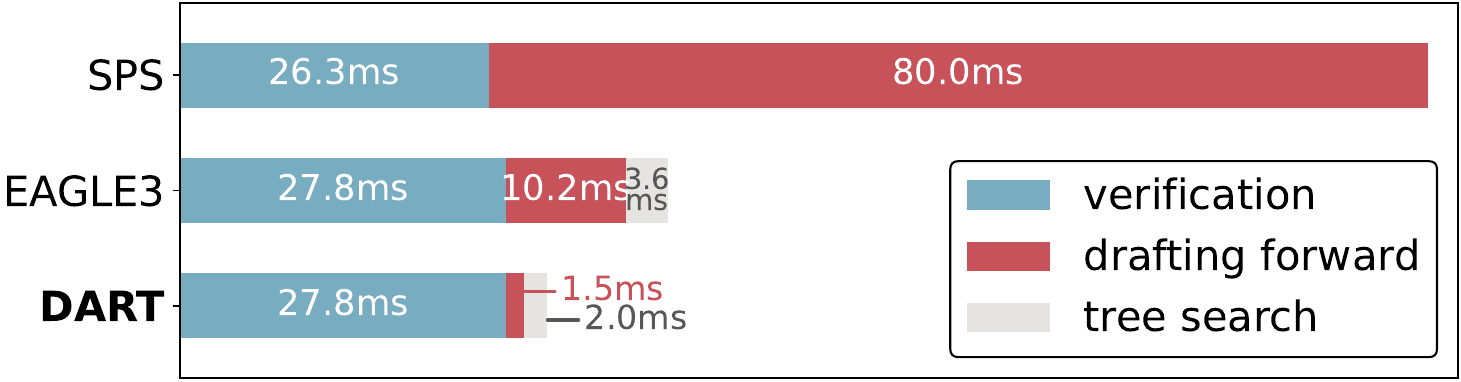}
  \caption{Latency of verification, drafting forward and tree search in one draft-verify iteration when accelerating Qwen3-14B using SPS, EAGLE3 and DART. SPS has slightly lower latency in verification because of fewer draft tokens than EAGLE3 and DART.}
  \label{fig:5}
\end{figure}

\paragraph{Implementation.} 
Unless otherwise specified, all experiments are conducted with a batch size of 1, and experiments on EAGLE3 reuse the weights from \citet{AngelSlim2025}. Other methods such as Medusa reuse their weights from their corresponding open-source repository. We do not compare DART with DiffuSpec or SpecDiff~\cite{christopher2025speculative} because their implementations are closed-source. 
Both DART and EAGLE3 use the draft length of 8.
We give the details of training and \textit{N}-gram-related settings in Appendix \ref{sec:appendix_train} and \ref{sec:appendix_ngram}.

\paragraph{Metrics.} In this paper, we focus on the following performance-related metrics. We do not report generation quality, since DART preserves the exact output distribution of the target model.
\begin{itemize}
  \item \textbf{Throughput.}
  The actual test end-to-end speedup ratio relative to vanilla autoregressive decoding.

  \item \textbf{Average acceptance length ($\tau$).}
  The average number of tokens generated per drafting-verification cycle, indicating the number of tokens accepted by the target model's verification.

  \item \textbf{Drafting latency.}
  The wall-clock time spent in the draft model forward and tree search to propose candidate tokens for each verification cycle. This metric captures the overhead introduced by the draft model and directly impacts the overall end-to-end speedup.
\end{itemize}

\subsection{Main Results}

\paragraph{Throughput improvement.}
We use vanilla autoregressive decoding as the baseline and compare DART against recent lossless speculative decoding methods, including standard speculative sampling (SPS), PLD, Hydra, Lookahead, Medusa and EAGLE3. As shown in Figure~\ref{fig:2} and Table~\ref{tab:performence}, DART achieves a 2.03$\times$--3.44$\times$ throughput improvement across tasks, outperforming EAGLE3 by 30\% on average. Notably, DART yields larger gains on code-related benchmarks. For example, in CodeAlpaca, DART surpasses EAGLE3 by 65\% when accelerating Qwen3-14B.

\paragraph{Relatively high $\tau$.}
As shown in Table~\ref{tab:performence}, DART achieves relatively high $\tau$ across all model scales. While DART does not always exceed EAGLE3 in $\tau$, the difference is within 0.2, and thus negligible. More importantly, due to its substantially lower drafting latency, DART converts a comparable $\tau$ into significantly higher throughput. This highlights that minimizing drafting overhead, rather than solely maximizing $\tau$, is critical for achieving superior end-to-end performance.

\paragraph{Drafting latency reduction.}
We focus on the drafting latency of 3 model-based drafters, including SPS, DART and EAGLE3. Methods like PLD and Lookahead do not have a separate model-based drafter, so their drafting latency is not reported. As shown in Figure~\ref{fig:5}, DART consumes 1.5ms latency to finish the drafting forward and reduces drafting forward latency by $6.8\times$ and $53.3\times$ compared to EAGLE3 and SPS in case of inference of Qwen3-14B. The tree pruning will take another 2ms to get the final draft token tree, which is also negligible latency and more efficient than the tree search of EAGLE3. This enables DART to accelerate LLM inference with negligible drafting overhead.

\subsection{Ablation Study}
\label{sec:ablation}

\paragraph{\textit{N}-gram effectiveness.}
The \textit{N}-gram constraint is a key component in DART for pruning the exponentially large search space induced by parallel token prediction. To evaluate its impact on drafting efficiency, we compare average acceptance length $\tau$ with and without \textit{N}-gram pruning across multiple benchmarks. As shown in Table~\ref{tab:ngram}, incorporating \textit{N}-gram pruning consistently leads to substantial improvements in $\tau$ across all evaluated benchmarks. This demonstrates that \textit{N}-gram pruning effectively removes low-quality branches in the draft token tree while preserving high-probability continuations, resulting in more efficient draft construction and higher acceptance rates during verification.

\begin{table}[t]
  \centering
  \caption{Effect of \textit{N}-gram pruning on drafting efficiency. Reported values are the average number of accepted draft tokens $\tau$ measured under speculative decoding on each benchmark.
}
  \label{tab:ngram}
  \resizebox{\linewidth}{!}{
    \begin{tabular}{c ccccc} 
    \toprule
    - & HumanEval & Alpaca & Math500 & CodeAlpaca \\
    \midrule
    w/o \textit{N}-gram  & 3.13 & 2.76 & 2.89 & 3.85\\
    \textbf{w/ \textit{N}-gram} 
    & \textbf{3.63} {\small\textcolor{gain}{($\uparrow$0.5)}}
    & \textbf{3.26} {\small\textcolor{gain}{($\uparrow$0.5)}}
    & \textbf{3.37} {\small\textcolor{gain}{($\uparrow$0.48)}}
    & \textbf{4.59} {\small\textcolor{gain}{($\uparrow$0.74)}}\\
    \bottomrule
    \end{tabular}
  }
\end{table}

\paragraph{Shifted logits prediction.}
We investigate the effect of shifted logits prediction in the DART inference pipeline. Specifically, we train two DART models based on Qwen3-4B from scratch under identical training configurations: one model adopts shifted logits prediction, while the other uses unshifted logits prediction. Unshifted logits prediction means logits predicted at
masked positions are interpreted as predictions for the tokens at the same positions. After training, we evaluate the prediction accuracy at multiple future positions on the same test data. The results are summarized in Table~\ref{tab:shifted}. We observe that shifted logits prediction substantially improves the accuracy of the first predicted position, increasing from 57.7\% to 71.1\%. In addition, modest but consistent accuracy gains are observed for subsequent positions in both hit@1 and hit@10 metrics.

\begin{table}[h]
  \centering
  \caption{Accuracy of multiple future positions under shifted and unshifted logits prediction. Hit@k denotes the proportion of cases where the ground-truth token appears among the top-$k$ predictions ranked by model logits. $\alpha$-$k$ denotes the prediction accuracy at the $k$-th future position.
  }
  \resizebox{\linewidth}{!}{
    \begin{tabular}{cccccc}
    \toprule
    Position  & $\alpha$-1  & $\alpha$-2  & $\alpha$-3 \\
    \midrule
    Unshifted 
      & 57.7\% & 38.1\% & 25.1\% \\
    \textbf{Shifted}
      & \textbf{71.1\%} {\small\textcolor{gain}{($\uparrow$13.4)}}
      & \textbf{41.0\%} {\small\textcolor{gain}{($\uparrow$2.9)}}
      & \textbf{27.6\%} {\small\textcolor{gain}{($\uparrow$2.5)}} \\
    \midrule
    Unshifted(hit@10)
      & 87.1\% & 74.2\% & 63.2\% \\
    \textbf{Shifted(hit@10)}
      & \textbf{93.2\%} {\small\textcolor{gain}{($\uparrow$6.1)}}
      & \textbf{76.7\%} {\small\textcolor{gain}{($\uparrow$2.5)}}
      & \textbf{65.9\%} {\small\textcolor{gain}{($\uparrow$2.7)}} \\
    \bottomrule
    \end{tabular}
  }
  \label{tab:shifted}
\end{table}

\paragraph{Annealing coefficient in DART training.}
During DART training, we apply annealed KL divergence objective to prevent the increasing uncertainty of longer-horizon predictions from disturbing overall training stability. Specifically, we introduce an annealing coefficient $\gamma$ to progressively downweight the supervision on future positions. To identify an appropriate value of $\gamma$, we train multiple DART models for Qwen3-4B from scratch with $\gamma \in \{0.5, 0.6, 0.7, 0.8, 0.9\}$, all using shifted logits prediction. The case $\gamma=1.0$ corresponds to no annealing and serves as the shifted-logits baseline from the previous ablation. Table~\ref{tab:anneal} compares the prediction accuracy at multiple future positions as well as the resulting average accepted length $\tau$. We observe a clear trade-off: smaller values of $\gamma$ improve accuracy at earlier positions while degrading performance at later positions, whereas larger $\gamma$ favors longer-horizon predictions. Since accuracy at early positions is more critical for speculative decoding, annealing the KL objective leads to a higher average acceptance length $\tau$, which translates linearly into end-to-end decoding throughput. Among all settings, $\gamma=0.6$ achieves the best balance between early-position accuracy and acceptance length $\tau$, and is therefore used as the default configuration in DART.

\begin{table}[t]
  \centering
  \caption{Effect of the annealing coefficient $\gamma$ on prediction accuracy and average acceptance length $\tau$ on HumanEval.
}
  \resizebox{\linewidth}{!}{
    \begin{tabular}{c cccccc c}
    \toprule
    $\gamma$ 
    & $\alpha$-1 
    & $\alpha$-2 
    & $\alpha$-3 
    & $\alpha$-4 
    & $\alpha$-5 
    & $\alpha$-6 
    & $\tau$ \\
    \midrule
    0.5 
      & 76.2\% & 45.0\% & 28.7\% & 18.4\% & 13.0\% & 10.6\% & 3.59 \\
    \textbf{0.6} 
      & \textbf{75.5\%} & \textbf{44.7\%} & \textbf{29.0\%} & \textbf{19.2\%} & \textbf{13.8\%} & \textbf{11.2\%} & \textbf{3.63} \\
    0.7 
      & 74.7\% & 44.1\% & 28.9\% & 19.6\% & 14.3\% & 11.8\% & 3.57 \\
    0.8 
      & 73.8\% & 43.2\% & 28.8\% & 19.8\% & 14.8\% & 12.2\% & 3.54 \\
    0.9 
      & 72.5\% & 42.1\% & 28.1\% & 19.4\% & 14.9\% & 12.4\% & 3.52 \\
    \midrule
    1.0 
      & 71.1\% & 41.0\% & 27.6\% & 19.3\% & 15.0\% & 12.6\% & 3.48 \\
    \bottomrule
    \end{tabular}
  }
  \label{tab:anneal}
\end{table}

\subsection{Larger Batch Sizes Study}
We also conduct experiments with batch sizes larger than 1 with DART and EAGLE3. As shown in Table~\ref{tab:larger_batch}, although throughput improvement decays as the batch size increases due to the more compute-bound brought by larger batch size, DART still gets larger gains than EAGLE3.

\begin{table}[h]
  \centering
  \caption{Throughput improvements under larger batch sizes on the HumanEval benchmark, evaluated on H20-3e device, for EAGLE3 and DART relative to autoregressive decoding (without speculative decoding) when target model is Qwen3-4B.}
  \resizebox{\linewidth}{!}{
    \begin{tabular}{c cccccccc}
    \toprule
    Batch Size 
    & 2 & 4 & 8 & 16 & 24 & 32 & 48 & 64 \\
    \midrule
    EAGLE3 
      & 1.84$\times$ & 1.69$\times$ & 1.48$\times$ & 1.32$\times$ & 1.29$\times$ & 1.28$\times$ & 1.24$\times$ & 1.22$\times$ \\
    \textbf{DART} 
      & \textbf{2.16$\times$} & \textbf{2.01$\times$} & \textbf{1.77$\times$} & \textbf{1.57$\times$} & \textbf{1.51$\times$} & \textbf{1.48$\times$} & \textbf{1.47$\times$} & \textbf{1.45$\times$} \\
    \bottomrule
    \end{tabular}
  }
  \label{tab:larger_batch}
\end{table}

\section{Conclusion}

We present \ours{}, a diffusion-inspired speculative decoding framework that enables fast and practical LLM inference. By predicting token distributions for multiple future positions in parallel within a single draft model forward pass, \ours{} eliminates the latency inherent in autoregressive drafting. To transform the parallel logits into coherent draft sequences, we introduce an efficient \textit{N}-gram-guided tree pruning strategy, which preserves semantic continuity without sacrificing parallelism. Experiments show that \ours{} substantially reduces draft-stage overhead and achieves significant end-to-end decoding speedups over prior speculative decoding methods. Overall, \ours{} demonstrates that parallel, distribution-level drafting combined with structured pruning, offers a viable and effective alternative to conventional autoregressive draft designs.



\section*{Impact Statement}

This paper proposes work that advances the field of speculative decoding and acceleration of LLMs. 


\bibliography{dart}
\bibliographystyle{icml2026}

\clearpage

\newpage
\appendix
\onecolumn

\section{EAGLE3 Latency Analysis}
\label{sec:appendix_eagle3_analysis}
\begin{figure*}[h]
  \centering
  \includegraphics[width=\textwidth]{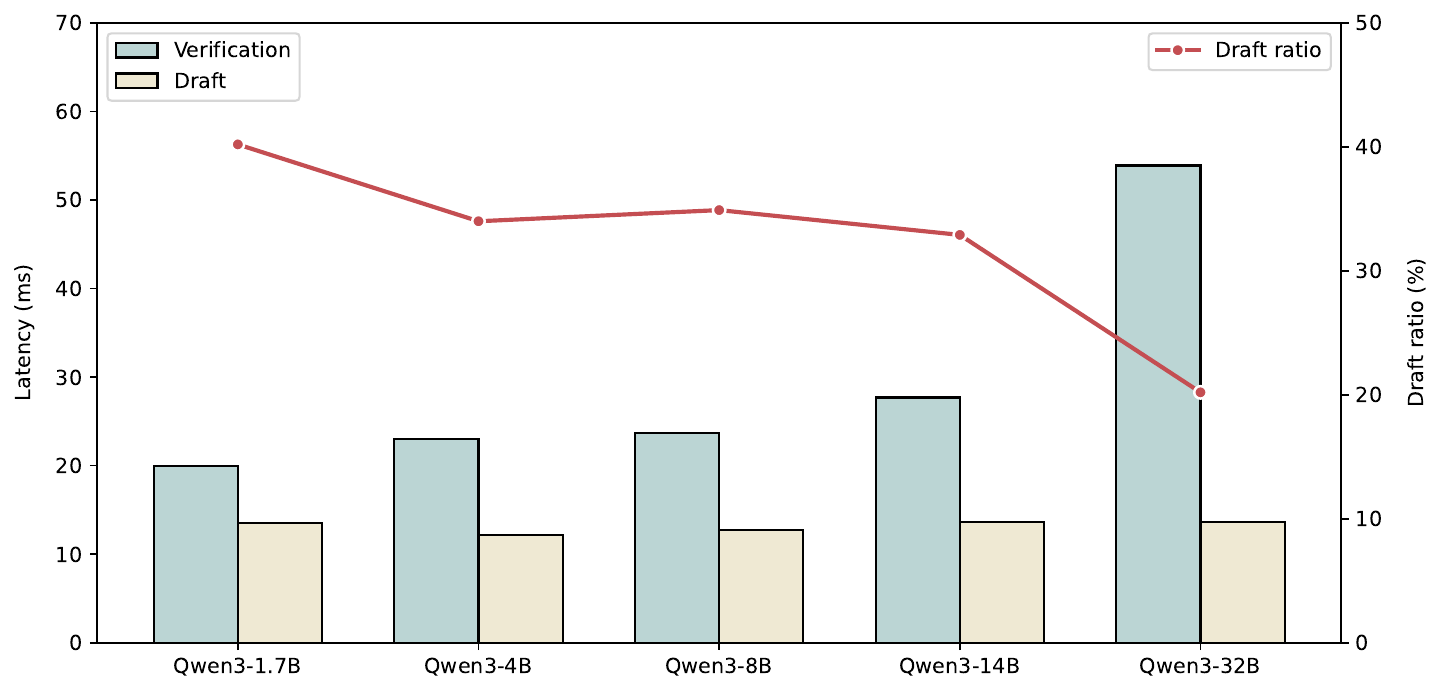}
  \caption{Per-round verification latency, drafting latency, and draft ratio across Qwen3 model variants evaluated on H20-3e. Bars (left axis) report per-round verification latency and drafting latency in milliseconds, while the line plot (right axis) shows the draft ratio (\%). Results are shown for Qwen3 models ranging from 1.7B to 32B parameters.}
  \label{fig:6}
\end{figure*}

EAGLE3's drafting process is inherently autoregressive. As shown in Figure \ref{fig:6}, EAGLE3's sequential dependency forces the drafter to spend nearly 20\%–40\% of the total inference time, thereby fundamentally limiting the achievable acceleration. 

Qwen3-32B exhibits a lower draft ratio compared to other models in the Qwen3 family. Although both Qwen3-32B and Qwen3-14B share a hidden dimension of 5120—resulting in nearly identical draft model parameters and latency within the EAGLE3 single-layer transformer framework—their verification overhead differs significantly. With 64 transformer layers compared to the 14B model's 40 \cite{qwen3}, Qwen3-32B incurs nearly double the verification latency. Because drafting time remains constant while verification costs scale with depth, the overall draft ratio for Qwen3-32B is inherently diminished.

\section{Training Details}
\label{sec:appendix_train}

\subsection{Training Implementation}
The code of training DART is based on SGLang's open-source repository SpecForge \cite{specforge2025}. 

\subsection{Training Setup}
We train the DART models of Qwen3-family and LLaMA2-Chat-7B under our carefully designed recipe. Unless otherwise specified, all training adopt a context length of 6400 and draft length of 8. We use the AdamW \cite{loshchilov2017decoupled} optimizer, with gradient clipping of 0.5 and beta values ($\beta_1$, $\beta_2$) set to (0.9, 0.999). The learning rate is set to 2e-5, and training is conducted for 3 epochs. DART is optimized using a cosine annealing learning rate schedule with linear warmup. 
The more detailed hyperparameters (e.g., number of heads, hidden dimension) of the customized Transformer decoder layer in DART keeps aligned with the target model.
Training is conducted on 8 $\times$ NVIDIA H20-3e GPUs.

\paragraph{Trainable parameters}
As shown in Figure~\ref{fig:3}, the trainable parameters in DART include the fully connected (FC) layer (whose output dimensionality matches the hidden size of the target model), decoder layer, and the LM head. 
DART shares the embedding layer with the target LLM, which remains frozen during training. 
In addition, the mask representation is also treated as a set of trainable parameters and is initialized with random values. This design choice follows the practice adopted in dLLM \cite{nie2025large}.

\paragraph{Training data.}  
Our training data is drawn from ShareGPT and UltraChat~\cite{ding2023enhancing}. 
After filtering, the combined dataset contains approximately 280K examples. 
We use the same data sources as EAGLE3, ensuring consistency in the training corpus.
DART takes the hidden states of the target model as input. 
Specifically, we select the outputs of the $1$st, 
$(\texttt{num\_layers}/2 - 1)$-th, and 
$(\texttt{num\_layers} - 4)$-th transformer layers as inputs to DART, 
following the same layer selection strategy as EAGLE3.

\paragraph{Training with Flex Attention.}  
We give the pseudo code of DART's sparse attention mask with Flex Attention.

\definecolor{codegreen}{rgb}{0,0.6,0}
\definecolor{codegray}{rgb}{0.5,0.5,0.5}
\definecolor{codepurple}{rgb}{0.58,0,0.82}
\definecolor{backcolour}{rgb}{0.95,0.95,0.92}

\lstdefinestyle{mystyle}{
    commentstyle=\itshape\color{red!50!green!50!blue!50},
    keywordstyle=\color{black},
    numberstyle=\tiny\color{codegray},
    stringstyle=\color{codepurple},
    basicstyle=\ttfamily\tiny,
    breaklines=true,
    captionpos=b,
    keepspaces=true,
    numbers=left,
    numbersep=5pt,
    showspaces=false,
    showstringspaces=false,
    showtabs=false,
    tabsize=2,
    frame=shadowbox,
}
\lstset{style=mystyle}

\begin{lstlisting}[language=Python]
def forward_flexattn_dart(
        hidden_states,          # hidden states of target LLM
        attention_mask,         # attention mask, where 1=valid, 0=pad
        position_ids,           # position ids of tree structure
        draft_len,              # draft length d
        rotary_emb,             # RoPE module (details omitted)
    ):
    B, q_len, _ = hidden_states.shape
    prompt_len = q_len // (draft_len + 1)
    Q, K, V = q_proj(hidden_states), k_proj(hidden_states), v_proj(hidden_states)
    Q, K = apply_rope(Q, K, rotary_emb, position_ids, q_len)
    key_cache, value_cache = K, V
    seq_lengths = attention_mask.sum(dim=-1)
    create_block_mask_func = compile_friendly_create_block_mask
    flex_attention_func    = compile_friendly_flex_attention
    mask_mod = generate_dart_mask(
        seq_lengths=seq_lengths,
        prompt_len=prompt_len,
        draft_len=draft_len,
    )
    block_mask = create_block_mask_func(
        mask_mod=mask_mod,    # function(b,h,q_idx,kv_idx)->bool
        B=B,
        H=1,                  # rely on broadcast over heads
        Q_LEN=q_len,
        KV_LEN=key_cache.shape[-2],
        device=Q.device,
    )
    attn_out = flex_attention_func(
        query=Q,
        key=key_cache.contiguous(),
        value=value_cache.contiguous(),
        block_mask=block_mask,
        enable_gqa=True,
    )
    out = o_proj(reshape_back(attn_out))   # -> (B, q_len, dim)
    return out
\end{lstlisting}

\begin{lstlisting}[language=Python]
def generate_dart_mask(seq_lengths, prompt_len, draft_len):
    def prompt_causal(b, h, q_idx, kv_idx):
        in_prompt = (q_idx < prompt_len) and (kv_idx < prompt_len)
        causal    = (q_idx >= kv_idx)
        valid     = (q_idx < seq_lengths[b]) and (kv_idx < seq_lengths[b])
        return in_prompt and causal and valid
    def draft_view_prompt(b, h, q_idx, kv_idx):
        is_draft      = (q_idx >= prompt_len)
        kv_in_prompt  = (kv_idx < prompt_len)
        draft_group = (q_idx - prompt_len) // draft_len
        valid_group = (draft_group < seq_lengths[b])
        allowed_prompt = (kv_idx <= draft_group)
        return is_draft and kv_in_prompt and valid_group and allowed_prompt
    def draft_internal_causal(b, h, q_idx, kv_idx):
        is_draft_q = (q_idx >= prompt_len)
        is_draft_k = (kv_idx >= prompt_len)
        q_group = (q_idx - prompt_len) // draft_len
        k_group = (kv_idx - prompt_len) // draft_len
        same_group = (q_group == k_group)
        causal     = (q_idx >= kv_idx)
        valid      = (q_group < seq_lengths[b])
        return is_draft_q and is_draft_k and same_group and causal and valid
    mask_mod = or_masks(prompt_causal, or_masks(draft_view_prompt, draft_internal_causal))
    return mask_mod
\end{lstlisting}

\section{\textit{N}-gram Details}
\label{sec:appendix_ngram}

\subsection{\textit{N}-gram Construction}
We implement the \textit{N}-gram using a trie data structure to achieve high retrieval performance. The construction relies on open-source dataset Dolma 3 Mix \cite{olmo2025olmo}, which covers multiple fields and types. The \textit{N}-gram is built based on the tokenizer of Qwen3 family or LLaMA2 family. To balance retrieval efficiency and memory overhead during inference, we adopt a \textit{3}-gram model in DART, which consists of approximately 1.3 billion tree nodes and occupies about 43.5 GB of disk space. 

\subsection{\textit{N}-gram Usage at Inference}
During inference, the \textit{N}-gram model is queried to provide scores for candidate tokens given the preceding context. As mentioned in Algorithm~\ref{alg:pipeline}, there are $k_i$ possible token candidates at position $i$ given a preceding context. To efficiently retrieve the scores for all $k_i$ candidate tokens with the same preceding context, we leverage the structural properties of trie. Specifically, we first locate the node corresponding to the preceding context. From this node, we can directly access all its child nodes, each representing a candidate token. This approach allows us to retrieve scores for all $k_i$ candidates in a single traversal, significantly reducing the number of required lookups compared to querying each candidate token individually. 

\begin{figure*}[h]
  \centering
  \includegraphics[width=\textwidth]{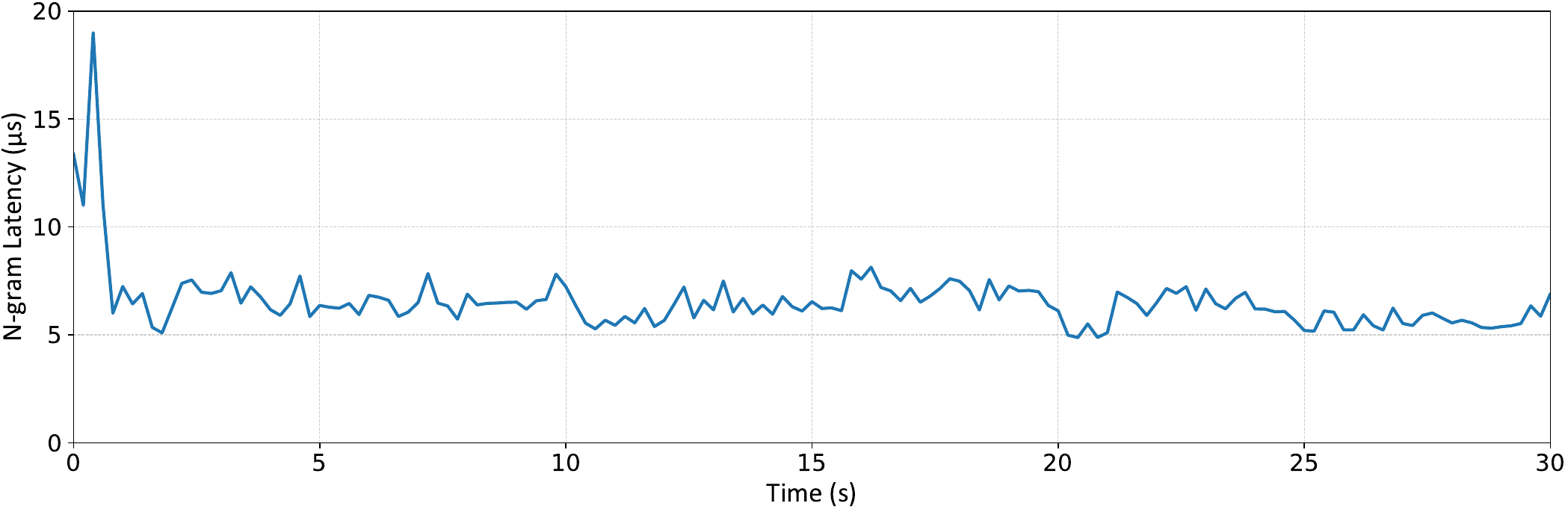}
  \caption{\textit{N}-gram retrieval latency over time. The timeline is divided into 200~ms intervals, and the average latency per query is recorded for each interval. After the warmup period, the average latency stabilizes around 6~$\mu$s per query after warmup.}
  \label{fig:ngram_latency_vs_time}
\end{figure*}

To support this efficient retrieval mechanism, we implement the \textit{N}-gram trie in \texttt{C++} and load it into CPU RAM before inference. During execution, the \textit{N}-gram trie occupies roughly 100~GB of CPU RAM. While this memory footprint may appear large, it is well aligned with modern inference servers, which typically provision hundreds of gigabytes of system memory and often leave substantial CPU RAM underutilized during LLM inference \cite{ma2025memory,luo2025headinfer,kong2024swapmoe}. Moreover, the footprint is shared across multiple inference processes. Thus, the increased RAM usage represents a practical design trade-off to achieve low-latency \textit{N}-gram retrieval, rather than a fundamental limitation of our approach. As shown in Figure~\ref{fig:ngram_latency_vs_time}, the average latency of \textit{N}-gram retrieval is around 6 $\mu$s per query after warmup, which is negligible compared to the overall inference time.

\section{Tree Pruning Algorithm Details}
\label{sec:appendix_algr}

This section provides the detailed hyperparameter settings and implementation specifics of the continuity-aware tree pruning algorithm described in Algorithm~\ref{alg:pipeline}. The pruning procedure is implemented in \texttt{C++} for efficiency, and parallelized using \texttt{OpenMP} along both the batch dimension and the candidate expansion process.

At each future position $i \in \{1, \ldots, d\}$, we select the top-$k_i$ tokens from the corresponding draft logits $l_i$. In all experiments, we set a uniform candidate threshold $k_i = 25$ for all positions. The active candidate set is maintained using beam pruning with beam width $w = 20$, which controls the maximum number of partial sequences retained at each depth.

The global draft token tree $\mathcal{T}$ is bounded by a maximum size $\theta = 59$, and only the top-$\theta$ scoring nodes are preserved for downstream verification. This constraint ensures a balance between drafting diversity and verification efficiency.

For each candidate extension, we compute a combined score consisting of a logit-based likelihood term and an \textit{N}-gram continuity score. Specifically, the logit score is defined as
\[
s_{\text{logit}} = \log(softmax(l)_{i,t} + \epsilon),
\]
where $l_{i,t}$ is the value of the logit for token $t$ at position $i$, and the \textit{N}-gram score $s_{\text{ng}}$ is defined as
\[
s_{\text{ng}} = \log(\mathrm{Pr}(t\ |\ \text{context}) + \epsilon),
\]
where $\mathrm{Pr}(t\ |\ \text{context})$ is the conditional probability of token $t$ provided by the pretrained \textit{N}-gram model, and $\epsilon$ is a small constant for numerical stability.

The final score increment is computed via the \textsc{Combine} function:
\[
\textsc{Combine}(s_{\text{logit}}, s_{\text{ng}})
= (w_{\text{logit}} \cdot s_{\text{logit}} + w_{\text{ng}} \cdot s_{\text{ng}}) \cdot w_{\text{level}},
\]
where the \textit{N}-gram weight is fixed to $w_{\text{ng}} = 0.5$, and the logit weight decays with tree depth (\textit{level} refers to the depth in the draft tree):
\[
w_{\text{logit}} = 0.9^{\text{level}}.
\]
This design prioritizes \textit{N}-gram continuity at deeper levels because of the low accuracy of logit of later positions. The level weight $w_{\text{level}}$ is defined as
\[
w_{\text{level}} = (\text{level} + 1)^{-0.7},
\]
which encourages longer draft sequences by favoring nodes at greater depths.

In practice, we implement the pruning algorithm in \texttt{C++} and parallelize the it with \texttt{OpenMP}. We also bind both \textit{N}-gram and worker threads to the same NUMA node to avoid expensive cross-node memory access overheads. 

Overall, the proposed pruning strategy efficiently constructs a compact yet high-quality draft token tree by jointly considering model confidence, token continuity, and structural constraints.

\begin{table}[t]
  \centering
  \small
  \setlength{\tabcolsep}{8pt}
  \renewcommand{\arraystretch}{1.15}
  \caption{Throughput and relative speedup across batch sizes evaluated on H20-3e (141G) device when target model is Qwen3-4B.}
  \label{tab:throughput_speedup_batchsize_qwen4b}

  \begin{tabular}{c c c c c c}
    \toprule
    & \multicolumn{3}{c}{Throughput (tokens/s)} & \multicolumn{2}{c}{Speedup} \\
    \cmidrule(lr){2-4}\cmidrule(lr){5-6}
    Batch size & Baseline & EAGLE3 & DART & EAGLE3 & DART \\
    \midrule
     2  & 105.8 & 194.6 & \textbf{228.7} & 1.84× & \textbf{2.16×} \\
     4  & 172.1 & 291.1 & \textbf{346.3} & 1.69× & \textbf{2.01×} \\
     8  & 243.3 & 359.0 & \textbf{431.1} & 1.48× & \textbf{1.77×} \\
    16  & 309.4 & 408.6 & \textbf{484.4} & 1.32× & \textbf{1.57×} \\
    24  & 338.3 & 436.4 & \textbf{510.9} & 1.29× & \textbf{1.51×} \\
    32  & 346.0 & 443.2 & \textbf{512.9} & 1.28× & \textbf{1.48×} \\
    48  & 364.6 & 450.9 & \textbf{538.5} & 1.24× & \textbf{1.47×} \\
    64  & 370.0 & 453.3 & \textbf{538.7} & 1.22× & \textbf{1.45×} \\
    \bottomrule
  \end{tabular}
\end{table}

\begin{table}[t]
  \centering
  \small
  \setlength{\tabcolsep}{8pt}
  \renewcommand{\arraystretch}{1.15}
  \caption{Throughput and relative speedup across batch sizes evaluated on H20-3e (141G) device when target model is Qwen3-8B.}
  \label{tab:throughput_speedup_batchsize_qwen8b}

  \begin{tabular}{c c c c c c}
    \toprule
    & \multicolumn{3}{c}{Throughput (tokens/s)} & \multicolumn{2}{c}{Speedup} \\
    \cmidrule(lr){2-4}\cmidrule(lr){5-6}
    Batch size & Baseline & EAGLE3 & DART & EAGLE3 & DART \\
    \midrule
     2  & 94.3  & 182.1 & \textbf{203.5} & 1.93× & \textbf{2.16×} \\
     4  & 160.6 & 246.5 & \textbf{268.6} & 1.53× & \textbf{1.67×} \\
     8  & 229.6 & 290.9 & \textbf{314.8} & 1.26× & \textbf{1.36×} \\
    16  & 297.2 & 327.8 & \textbf{343.3} & 1.09× & \textbf{1.14×} \\
    24  & 329.4 & 335.7 & \textbf{353.5} & 1.01× & \textbf{1.06×} \\
    32  & 339.7 & 351.4 & \textbf{353.6} & 1.02× & \textbf{1.04×} \\
    48  & 354.2 & 352.9 & \textbf{366.4} & 0.99× & \textbf{1.03×} \\
    64  & 360.3 & 356.4 & \textbf{366.1} & 0.98× & \textbf{1.01×} \\
    \bottomrule
  \end{tabular}
\end{table}

\section{Larger Batch sizes Study Details}
We provide additional results on the performance of DART and EAGLE3 under larger batch sizes. Specifically, we report throughput and relative speedup for EAGLE3 and DART on Qwen3-4B and Qwen3-8B, as summarized in Tables~\ref{tab:throughput_speedup_batchsize_qwen4b} and~\ref{tab:throughput_speedup_batchsize_qwen8b}.

\section{DART Performance in A100}
Given that speculative decoding is sensitive to hardware-specific balances between memory and compute, we further explore DART's cross-platform generalizability on NVIDIA A100-40G GPU. Under evaluation using a single A100-40G GPU, DART exhibits strong and consistent performance across the Qwen3 family (1.7B, 4B, 8B, and 14B). Specifically, DART achieves roughly 30\% speedup over EAGLE3 (see Table \ref{tab:performence_a100}), highlighting the hardware-agnostic efficiency of our proposed design.

\begin{table*}[t]
  \centering
  \caption{Speedup ratios of different methods and mean average acceptance lengths $\tau$ on MT-Bench, HumanEval, Alpaca, Math500, CodeAlpaca, LiveCodeBench and MBPP, evaluated on a single NVIDIA A100 GPU. All Qwen models are from Qwen3 family, for example, Qwen4B represents Qwen3-4B.}
  \label{tab:performence_a100}
  \newcommand{\cb}{\cellcolor{dartblue}}
  \resizebox{\linewidth}{!}{
    \begin{tabular}{ccccccccccc}
    \toprule
    \multirow{2}{*}{Model} &
    \multirow{2}{*}{Method} &
    \multicolumn{7}{c}{Speedup} &
    \multicolumn{2}{c}{Mean} \\
    
    \cmidrule(lr){3-9}
    \cmidrule(lr){10-11}
    
    &
    & Alpaca & Codealpaca & Humaneval & LiveCode & Math500 & MBPP & MT-bench
    & Speedup & $\tau$ \\
    
    \midrule

    \multicolumn{10}{c}{Temperature=0} \\
    \midrule

    \multirow{2}{*}{Qwen1.7B}
      & EAGLE3     & 1.91× & 2.26× & 2.32× & 2.28× & 2.46× & 2.19× & 2.09× & 2.21× & \textbf{3.80} \\
      & \cb\textbf{DART}
                    & \cb\textbf{2.31×}
                    & \cb\textbf{3.12×}
                    & \cb\textbf{2.88×}
                    & \cb\textbf{2.57×}
                    & \cb\textbf{2.94×}
                    & \cb\textbf{2.82×}
                    & \cb\textbf{2.50×}
                    & \cb\textbf{2.74×} & \cb3.60 \\
    \midrule

    \multirow{2}{*}{Qwen4B}
      & EAGLE3     & 2.20× & 2.29× & 2.29× & 2.16× & 2.21× & 2.17× & 2.23× & 2.22× & 3.54 \\
      & \cb\textbf{DART}
                    & \cb\textbf{2.44×}
                    & \cb\textbf{3.61×}
                    & \cb\textbf{2.83×}
                    & \cb\textbf{2.69×}
                    & \cb\textbf{2.57×}
                    & \cb\textbf{3.23×}
                    & \cb\textbf{2.42×}
                    & \cb\textbf{2.83×} & \cb\textbf{3.87} \\
    \midrule

    \multirow{2}{*}{Qwen8B}
      & EAGLE3     & 2.16× & 2.49× & 2.46× & 2.24× & 2.49× & 2.40× & 2.10× & 2.33× & \textbf{3.72} \\
      & \cb\textbf{DART}
                    & \cb\textbf{2.52×}
                    & \cb\textbf{3.50×}
                    & \cb\textbf{3.07×}
                    & \cb\textbf{2.84×}
                    & \cb\textbf{2.60×}
                    & \cb\textbf{3.23×}
                    & \cb\textbf{2.06×}
                    & \cb\textbf{2.83×} & \cb3.61 \\
    \midrule

    \multirow{2}{*}{Qwen14B}
      & EAGLE3     & 1.76× & 2.19× & 2.40× & 2.32× & 2.54× & 2.14× & 2.01× & 2.19× & 3.48 \\
      & \cb\textbf{DART}
                    & \cb\textbf{2.61×}
                    & \cb\textbf{3.40×}
                    & \cb\textbf{2.80×}
                    & \cb\textbf{2.83×}
                    & \cb\textbf{2.60×}
                    & \cb\textbf{3.21×}
                    & \cb\textbf{2.47×}
                    & \cb\textbf{2.84×} & \cb\textbf{3.67} \\
    \bottomrule
    \end{tabular}
  }
\end{table*}

\section{LLM Usage}
Large language models were used only for limited editorial assistance, such as improving grammar and clarity. They did not contribute to research ideas, experimental design, analysis, or conclusions. All content was written and verified by the authors, who take full responsibility for the accuracy of the manuscript.

\end{document}